\documentclass{ieeeaccess}
\usepackage{cite}
\usepackage{amsmath,amssymb,amsfonts}
\usepackage{algorithmic}
\usepackage{graphicx}
\usepackage{textcomp}

\usepackage{bm}
\makeatletter
\AtBeginDocument{\DeclareMathVersion{bold}
\SetSymbolFont{operators}{bold}{T1}{times}{b}{n}
\SetSymbolFont{NewLetters}{bold}{T1}{times}{b}{it}
\SetMathAlphabet{\mathrm}{bold}{T1}{times}{b}{n}
\SetMathAlphabet{\mathit}{bold}{T1}{times}{b}{it}
\SetMathAlphabet{\mathbf}{bold}{T1}{times}{b}{n}
\SetMathAlphabet{\mathtt}{bold}{OT1}{pcr}{b}{n}
\SetSymbolFont{symbols}{bold}{OMS}{cmsy}{b}{n}
\renewcommand\boldmath{\@nomath\boldmath\mathversion{bold}}}
\makeatother

\def\BibTeX{{\rm B\kern-.05em{\sc i\kern-.025em b}\kern-.08em
    T\kern-.1667em\lower.7ex\hbox{E}\kern-.125emX}}

\begin{document}
\history{Date of publication xxxx 00, 0000, date of current version xxxx 00, 0000.}
\doi{preprint}

\title{A Vision-Enabled Prosthetic Hand for Children with Upper Limb Disabilities}

\author{
\uppercase{Md Abdul Baset Sarker}\authorrefmark{1},
\uppercase{Art Nguyen}\authorrefmark{2}, 
\uppercase{Sigmond Kukla}\authorrefmark{3},
\uppercase{Kevin Fite}\authorrefmark{4},
\uppercase{Masudul H. Imtiaz}\authorrefmark{5}
}

\address[1]{Md Abdul Baset Sarker is with Clarkson University, Potsdam, NY-13699, USA (e-mail: sarkerm@clarkson.edu). }
\address[2]{Art Nguyen is with Clarkson University, Potsdam, NY-13699, USA (e-mail: nguyenqp@clarkson.edu). }
\address[3]{Sigmond Kukla is with Clarkson University, Potsdam, NY-13699, USA (e-mail: kuklasj@clarkson.edu). }
\address[4]{ Kevin Fite is with Clarkson University, Potsdam, NY-13699, USA (e-mail: kfite@clarkson.edu). }
\address[5]{ Masudul H. Imtiaz is with Clarkson University, Potsdam, NY-13699, USA (e-mail: mimtiaz@clarkson.edu). }

\tfootnote{Paper submission date Mar 31, 2025' This work was supported in part by Clarkson University, Potsdam, NY}

\markboth
{Author \headeretal: Preparation of Papers for IEEE TRANSACTIONS and JOURNALS}
{Author \headeretal: Preparation of Papers for IEEE TRANSACTIONS and JOURNALS}

\corresp{Corresponding author: Masudul H. Imtiaz (e-mail: mimtiaz@clarkson.edu).}

\begin{abstract}
This paper introduces a novel AI vision-enabled pediatric prosthetic hand designed to assist children aged 10–12 with upper limb disabilities. The prosthesis features an anthropomorphic appearance, multi-articulating functionality, and a lightweight design that mimics a natural hand, making it both accessible and affordable for low-income families. Using 3D printing technology and integrating advanced machine vision, sensing, and embedded computing, the prosthetic hand offers a low-cost, customizable solution that addresses the limitations of current myoelectric prostheses.  A micro camera is interfaced with a low-power FPGA for real-time object detection and assists with precise grasping. The onboard DL-based object detection and grasp classification models achieved accuracies of 96\% and 100\% respectively. In the force prediction, the mean absolute error was found to be 0.018. The features of the proposed prosthetic hand can thus be summarized as: a) a wrist-mounted micro camera for artificial sensing, enabling a wide range of hand-based tasks; b) real-time object detection and distance estimation for precise grasping; and c) ultra-low-power operation that delivers high performance within constrained power and resource limits.
\end{abstract}



\maketitle

\section{Introduction}
\label{sec:introduction}
\IEEEPARstart{C}{ongenital} limb loss and upper extremity abnormalities were estimated to occur in approximately 15 individuals per 100,000 live births in the United States alone \cite{cdc2019facts,behrend2011advances}. Beyond congenital disabilities, tumors, severe infections, or traumatic injuries also cause pediatric limb deficiency and place a significant physical and emotional burden on a child and their family. Replacement of an upper limb with a functional prosthetic hand had the potential to restore some limb functionality and improve the independence of these children. Furthermore, the earlier children were fitted for a powered prosthesis, the lower the rate of prosthetic hand rejection in the later years of their life \cite{toda2015use}. The challenges in fitting actuation and control systems in a small size while maintaining a comfortable weight for long-term use were among the main reasons for the limited options of hand prostheses for children. This paper addresses these limitations and presents a next-generation pediatric prosthetic hand with advanced control.

\begin{figure}[!ht]
\centering
\includegraphics[width=\columnwidth]{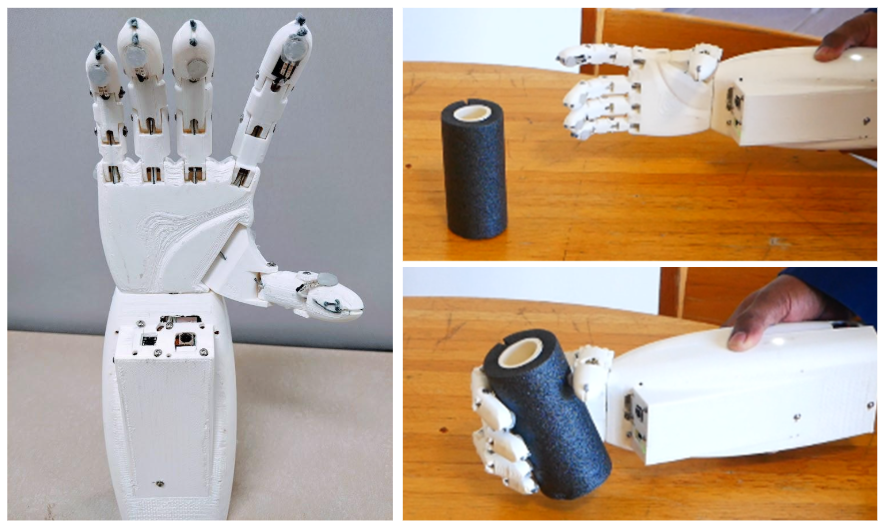}
\caption{Example of 3D printed vision-enabled prosthetic hand demonstrating grasping cylindrical object \cite{BasetSarker2022}.}
\label{prev_hand_work}
\end{figure}

The amalgamation of computer vision with different technologies has been explored in recent years \cite{alfaris2020review, ghofrani2019machine}, and vision-based control was successfully implemented in many research projects \cite{shi2020,deGol2016}. The motivation for this research came from our preliminary in-lab prosthesis designs (shown in Figure \ref{prev_hand_work}) for grown-ups \cite{BasetSarker2022} that used vision-based control to detect and grasp the object and a flexible pressure sensor to get tactile feedback. Before this approach, we also experimented with controlling hand gestures through brain waves \cite{ketola2022lessons}; however, acute user attention is required for this, and daily use is not feasible. The revised design \cite{BasetSarker2022} instead used a 2MP camera, a time-of-flight (TOF) distance sensor, and embedded processors to perform most of the grasping tasks. A per-finger pressure sensor provided feedback to ensure a safe grip, while an IMU detected gestures and released the object (video demonstration of this hand is in \cite{youtube_visioncontrolled_prosthetic}). Unlike EEG or EMG-based designs, vision-controlled hands require minimal personalized training. Each person needs to be trained to adapt EEG / EMG-controlled devices and personalized data is required to provide. Passive automation through a vision-based prosthesis may keep user input to a minimum, making it easier to use.  

The success of this research motivated us to improve the design by lowering the cost and power budget while adding edge computing technologies. The target population also changes in this project to elementary school-age children as their physical, social, and mental skills develop at this age \cite{weiss1991psychological}. The study shows that replacing an upper limb with a functional prosthetic hand can potentially return some of the functionality of the limb and improve the independence of these children. However, the majority of prostheses available to children were myoelectric-based \cite{cordella2016literature,ccorimanya2019myoelectric}, priced around 14,000 USD. Although myoelectric-based approaches have shown their limitations \cite{weiner2018kit}, other technologies had not been thoroughly investigated for commercialization. Hence, we aimed to present an alternative (vision-controlled) but smart prosthetic hand that is customizable for children. At the same time, we sought to keep costs and the power budget low to maintain accessibility.

The use of 3D printing in the fabrication of soft robotic systems \cite{laschi2016soft} allows the manufacturing of customized products at low volumes in a cost-effective way, particularly in the fabrication of hand prostheses. The soft structure of the prosthesis may provide a safer mechanism than conventional rigid-material systems. An anthropomorphic soft robotic prosthetic hand (X-Limb \cite{mohammadi2019xlimb}) and its revision for children \cite{mohammadi2020paediatric} were designed in a monolithic fashion, eliminating the need for assembly and the associated inadvertent misalignments. Our proposed design was in-house 3D-printed using a combination of hard PLA (polylactic acid) and soft material (silicone). The system (design, interface, and software efficiency) was fine-tuned (e.g., design refinement, component selection, computation management, and cost reduction) through successive iterations. 

In summary, with the overall goal of assisting younger children with upper-limb disabilities,  we leveraged soft prosthesis design and integrated machine vision, advanced sensing, and embedded computing into a prosthetic hand. This prosthesis had an anthropomorphic appearance, soft structure, multi-articulating functionality for grasping a wide range of objects, and low weight, with a size similar to the natural hand of the target population: children aged 10-12 years.  The technical contributions are as follows: 
\begin{enumerate}
\item presentation of a low-power FPGA-based digital design interfacing with a 2MP camera, 
\item development of a customized DL force and grasp classification in conjunction with an object detection model to be hosted on the FPGA and multiple sensor interface for firm grasping. 
\item presentation of the ultra-low-power control mechanism to ensure high performance within a limited resource, size, and power budget of the pediatric prosthesis.
\end{enumerate}

\section{The Big Picture}

As the project target was a subset of children (10-12 years old) with trans-radial amputations, the size of the developed prosthetic hand and forearm was designed to be similar to the biological hand of this group, and the weight was kept lower than that of the physiological limb. The average hand and forearm weight for 10-12-year-old children were 120 g and 320 g, respectively \cite{krishnan2016}. The average forearm-hand length for these children was 30 cm, hand length was 13 cm, and hand breadth was 6 cm \cite{snyder1977}. 

Typical prosthetic hands fall into two categories: fully actuated and under-actuated. A fully actuated prosthesis closely replicates the actions and mobility of a natural hand, including up to 27 Degrees of Freedom (DOF). However, to reduce complexity, cost, power consumption, and weight, we developed the prosthetic hand with an underactuated design. To meet the basic dexterity needs of this group, the design only incorporated essential grasp types: the power grasp (cylindrical and spherical) and pinch/tripod grasp as these were identified to cover more than 70\% of daily activities \cite{light2002_53}. Hence, a minimum of three degrees of actuation were included: one for the thumb, one for the index and middle fingers, and one for the ring and little fingers. 

The design target was to deliver the force of the finger up to 5 N, and the closing and opening target time is approximately 1.5 seconds. 

\section{Methodology}
\begin{figure*}[!t]
\centering
\centerline{\includegraphics[width=\textwidth,height=6cm,keepaspectratio]{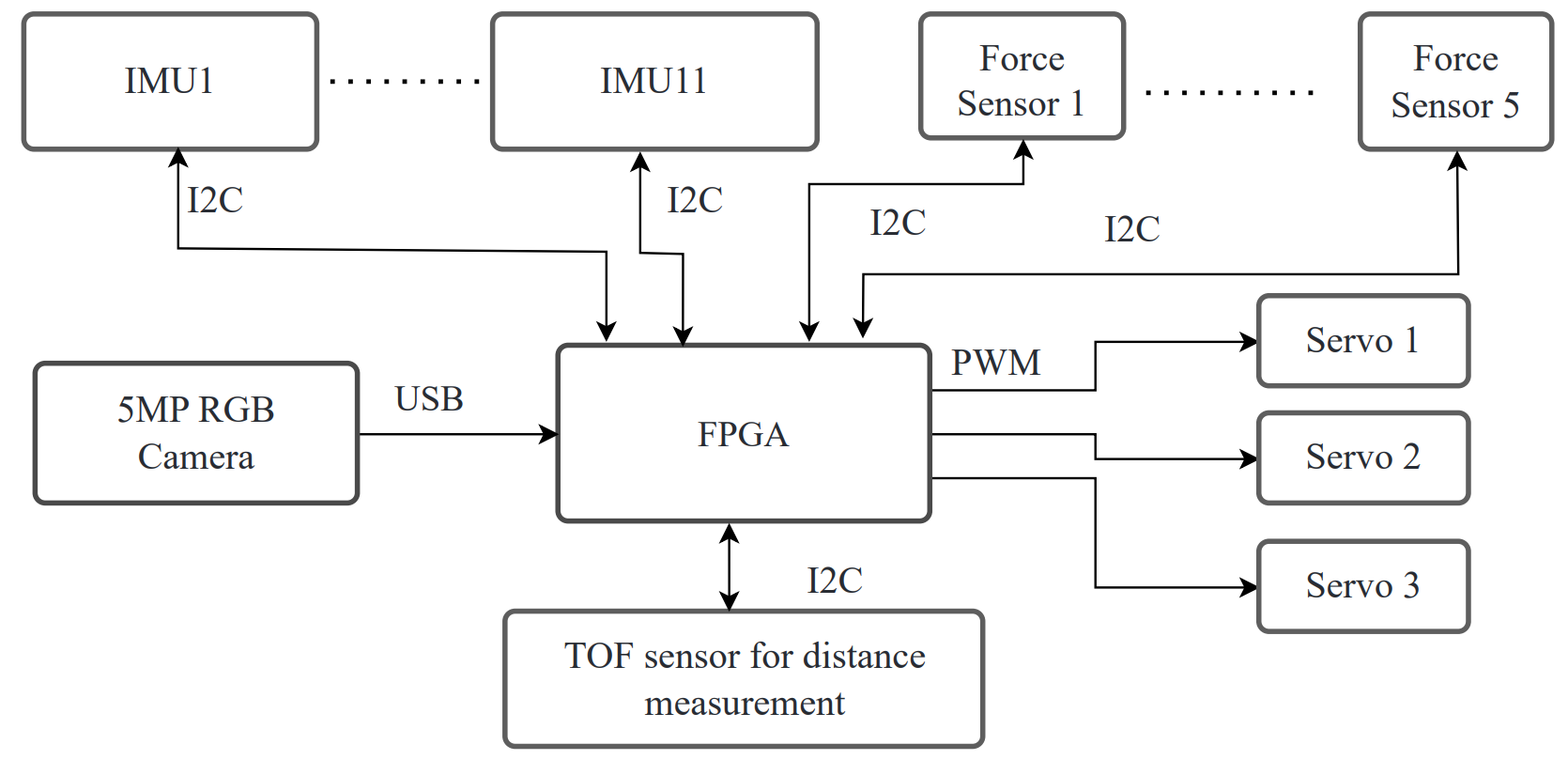}}
\caption{Block Diagram of Sensors and electronic connections with FPGA}
\label{block_diagram}
\end{figure*}

The proposed pediatric hand integrates multiple sensors and control components with an FPGA. The FPGA acts as the central processing unit, interfacing with several peripherals via different communication protocols (shown in Figure \ref{block_diagram}). It connects to 11 IMUs and 5 Force Sensors through the I2C interface to gather data related to motion and force. A 2MP RGB camera communicates with the FPGA via the USB interface, enabling image capture. Additionally, a TOF (Time-of-Flight) sensor is integrated using I2C for distance measurement. The FPGA processes all incoming data and controls the motors depending on the pressure feedback, thus controlling the movement of the fingers. 

Initially, the system remains in an idle state, waiting for a characteristic user gesture to be activated. Upon activation, the camera captures frames to detect nearby objects. Once an object is detected, the system selects the appropriate grasp pattern and force using the force and grasp classification model. If the object is within reach, the hand moves to grasp it. 
During grasping, the system monitors the force (a preset threshold in the grasp and object detection model) to ensure that the object is securely held and does not break. After grasping, the system waits for a characteristic tilt gesture to signal the release of the object, completing the task. The process repeats as needed, starting from gesture classification.

\subsection{Mechanical Design}
\begin{figure}[h]
\centering
\includegraphics[height=7cm]{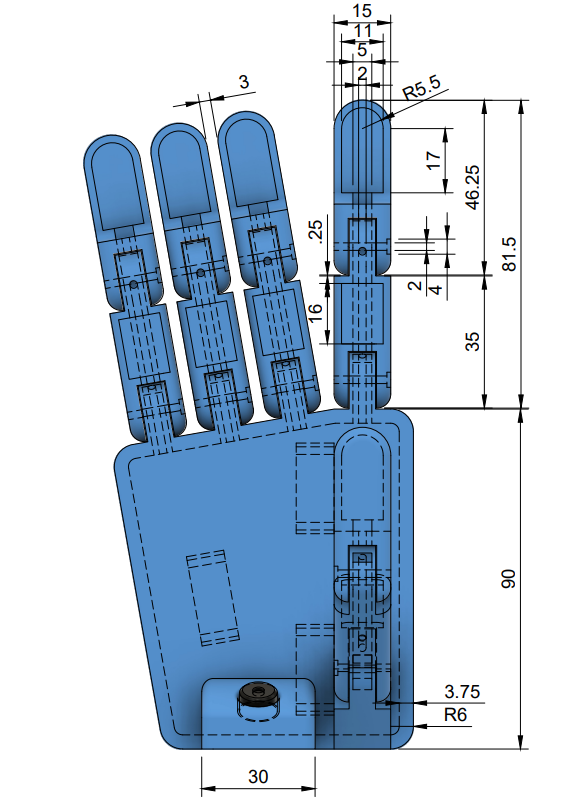}
\caption{Measurements are shown in the mechanical design of the hand. }
\label{hand_measurements}
\end{figure}

This prosthetic hand is specifically designed for children aged 10 to 12, carefully considering factors such as weight, usability, and functionality.  Figure \ref{hand_measurements} shows the measurements of the designed hand. The data on the size of the hand were obtained from \cite{pexoexo2019}.  To ensure light weight and convenience, we chose the underactuated design with 3 DOF, which balances sufficient dexterity with the simplicity required for reliable, everyday operation. Each finger is designed with two joints, as similar implementations found in robotic and prosthetic research \cite{stavenuiter2017,li2017novel,weiner2018kit,weinerFingerDesign}, where two joint structures provided effective adaptability and grasping versatility with minimal complexity. Fingers and structural components are made using PLA and 3D printing technology, which ensures affordability, lightweight properties, and ease of replacement. 

A soft silicone layer was added to the fingertips to improve grip and facilitate interaction with various objects. The silicone was prepared using a two-part liquid silicone rubber (LSR) system, poured into custom molds, and cured at room temperature, resulting in a durable, skin-like surface. The chosen silicone mixture, Smooth-On (80\% silicone base and 20\% curing agent), was determined through experimental testing to achieve the desired balance between softness and durability. When combined with the underlying PLA structure, this silicone layer enhances the contact area, thereby improving grasp performance during everyday tasks such as holding utensils or toys. Additionally, the silicone provides waterproof protection, safeguarding internal electronic components from moisture exposure during typical daily activities. 

\subsection{Electronic Design}
\subsubsection{FPGA}
In this design, a 2MP USB camera was interfaced with a Xilinx UltraScale+™ MPSoC processor, which combined a high-performance ARM processor and FPGA fabric, including 256K system logic cells, 1.2K DSP slices, and built-in image sensor processing (ISP) in the same chip.  The development was initiated on a Kria KV260 board; then, a customized PCB was developed. A CNN-based gesture classification, Force and grasp classification, and object detection model was hosted on this processor. 

\subsubsection{System Firmware}
For power-saving purposes, all modules were in sleep mode when the device was not applied to an amputee, sensed by the hand IMU. When applied, the camera and other sensors remained in default sleep mode, except the IMU. After detecting characteristic gestures from the IMU, the hand was activated with an interrupt to the FPGA processor to initialize the camera (within 0.3 seconds). When activated and moved close to an object, the camera captured the image and identified the type of object based on the hosted DL model. The controller got the object coordinates from the model and the distance from the sensor. When the system detected multiple objects closer to the hand, the system determined the hierarchy depending on the object bounding box size. If the system selects the wrong object, the user can correct it with a specific gesture. The fingertip pressure sensor and IMU ensured a precise grasp. When the user wanted to release the object, the characteristic gesture was detected by the IMU and sent to the FPGA.

\subsubsection{Fingertip sensors}
Each finger contains one force sensor (FMAMSDXX015WC2C3) and two IMUs (MPU-6500). A TOF (VL6180) distance sensor was placed on the index finger at the tip. We empirically found that for smaller objects, placing a distance sensor on the index finger works better than putting it on the wrist position. 

\begin{figure}[!h]
\centering
\includegraphics[width=\columnwidth]{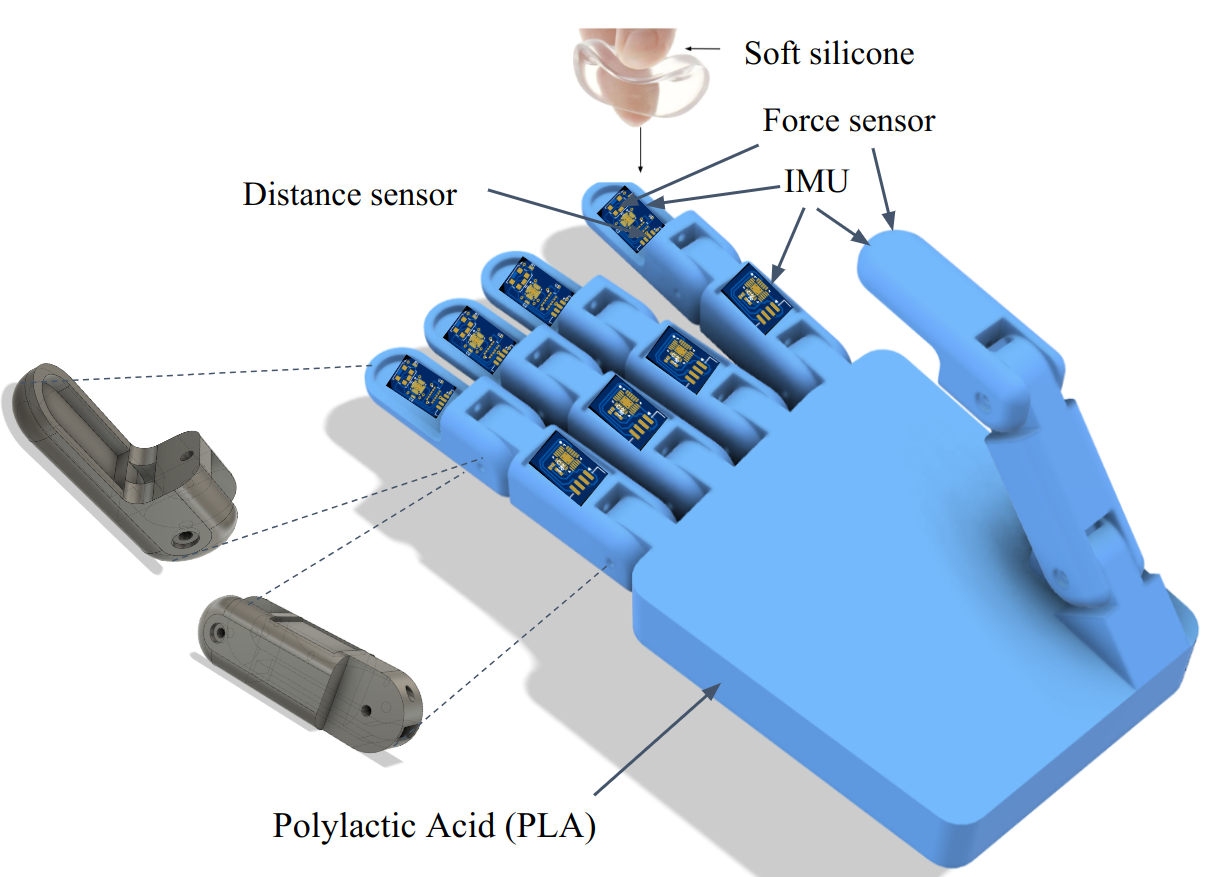}
\caption{Sensor placement on the finger.}
\label{finger_sensor_placement}
\end{figure}

\subsubsection{Finger Control System}
In this design, fingers are driven by the tendon wires connected to the servo motor (INJS2065 Servo, weight 20g, stall torque 7kg/cm).  The index and thumb fingers are connected to two servo motors, respectively. The rest of the fingers are connected to a one-servo motor. The number of motors is reduced to three to reduce power consumption and weight and build a small form factor. The input signal to run the motors comes from the FPGA. 

\subsection{Object detection model}
\label{object_detection_det}
YOLOv7-tiny is designed as an efficient, real-time object detector tailored for resource-constrained environments derived from (You Only Look Once) YOLOv7  \cite{wang2023yolov7}. It incorporates innovations from the full-scale YOLOv7 architecture—such as Extended Efficient Layer Aggregation Networks (E‑ELAN) and compound scaling strategies—to enhance gradient flow and optimize parameter usage, all while significantly reducing computational complexity. Compared to previous tiny models like YOLOv4-tiny, YOLOv7-tiny achieves higher detection accuracy and faster inference speeds, making it well-suited for deployment on edge devices where low latency and efficiency are critical. In this project, we planned to implement the model in KV260. To run the model into KV260 Deep Learning Processor Unit (DPU), we need to quantize the model. Different approaches exist, such as Quant aware training (QAT) and post-quantization. We chose QAT because it gives us slightly more accuracy. 

\begin{figure*}[!h]
\centering
\centerline{\includegraphics[width=\textwidth,height=9cm,keepaspectratio]{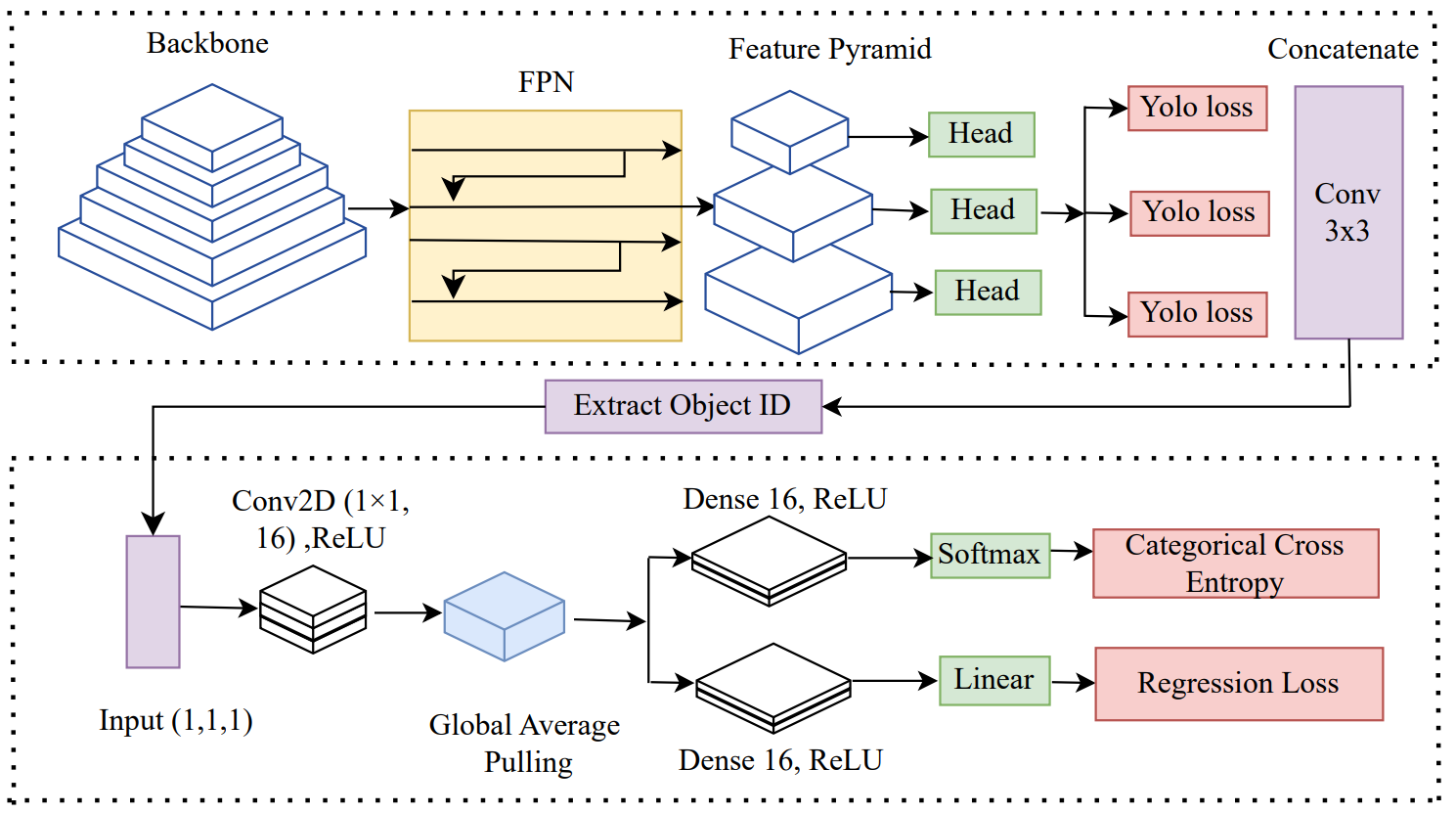}}
\label{object_detection_force_model}
\caption{ (Top) Object detection model \cite{wang2023yolov7} and (bottom) Force and Grasp Classification model.}
\end{figure*}

\subsubsection{Data collection and training object detection model}
\label{sec:ObjDetectionTraining}
\begin{figure}[!h]
\centering
\includegraphics[width=\columnwidth]{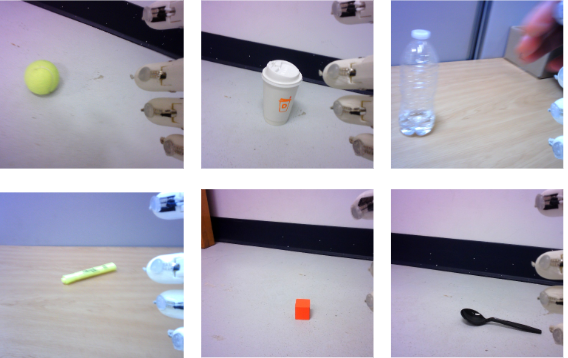}
\caption{Sample of data collected for training  object detection model.}
\label{some_example_of_collected_data}
\end{figure}

For model training, initially, images were captured from the 2MP camera placed on the wrist position for six object classes (ball, cup, bottle, pen, spoon, cube). Images are collected from different lighting conditions and different backgrounds. Data is collected from various distances from the object. Some sample images are shown in the Figure \ref{some_example_of_collected_data}. Data were divided into 80\%  training subsets, 10\%  testing, and 10\% validation subsets. Trained YOLOv7-tiny with 500 epochs using images size 640 pixels. Hyperparameter optimization resulted in parameters such as an initial learning rate (lr0) of 0.0105, a final learning rate factor (lrf) of 0.01, a momentum of 0.908, and a weight decay of 0.00041.

For data augmentation, we applied flipping and mixed augmentations that applied horizontal flips (50\% probability), doubling the orientations seen for each object. This helps the model handle objects viewed from either the left or right, improving viewpoint invariance. More significantly, YOLOv7-tiny leverages mosaic augmentation (enabled at 100\% frequency), which combines four images in one during training. Mosaic presents the model with cluttered scenes of multiple scaled objects and varied backgrounds, a technique credited with improving the detection of small and occluded objects.
For color augmentation, we have used (hsv\_h=0.015, hsv\_s=0.7, hsv\_v=0.4) to randomly shift the hue (±1.5\%), saturation (up to 70\%), and brightness (up to 40\%) of training images. This jittering creates diverse color and illumination conditions, which prevents the model from overfitting to a specific lighting scenario. As a result, the detector becomes more invariant to lighting changes and color variations. We have also used random geometric transformations like translation (±20\% shift) and scale (up to ~90\% resize). These enhancements teach the model to detect targets despite changes in position or distance, making it less sensitive to the exact location or scale of the object.

\subsection{Force and Grasp Classification}
\label{subsec:Force_and_Grasp_Classification}
The proposed model (Figure \ref{object_detection_force_model}) is designed for finger force prediction and grasp pattern classification based on objects detected by the object detection model. Initially, the extracted object ID is passed to the model by reshaping (1x1x1), then a Conv2D(1×1) filter size 16 with activation function RelU, followed by global average pooling. The resulting features feed into two parallel dense layers, each with 16 neurons and ReLU activation. One dense branch predicts the grasping pattern through softmax activation and categorical cross-entropy loss, while the other branch estimates the maximum grasping force using linear activation with regression loss. As the Grasping pattern is categorical and the force is a floating point number, we calculated categorical cross-entropy loss for Grasp classification and Mean Absolute Error (MAE) for Force.

\subsubsection{Force and Grasp Classification Training}
The model was trained using a data set comprising 3,000 pressure data points collected from a force sensor placed on the finger for three grasping patterns: power grip, pinch, and pronated. Data were gathered from six distinct objects of real-world situations, encompassing things of various sizes, shapes, and textures. The pressure sensors measured the force exerted by each finger during every grip, offering information on how the prosthetic hand emulates the natural movements of the human hand when manipulating various objects.  The data were split into training (80\%), testing (10\%), and validation (10\%). Training with a learning rate of 0.002 for 200 epochs, we achieved better performance.

\subsection{Gesture Classification}
The camera in the prosthetic hand is placed on the wrist. While grasping the object, the camera view is generally blocked, so we need another signal or input to enable the release of the object. To address this, we have added an IMU in the palm to detect the characteristic gesture. Upon detecting the gesture, the object is released. To train this specific model, we collected data using our developed prosthetic hand. 

\begin{figure*}[!ht]
\centering
\centerline{\includegraphics[width=\textwidth,height=6cm,keepaspectratio]{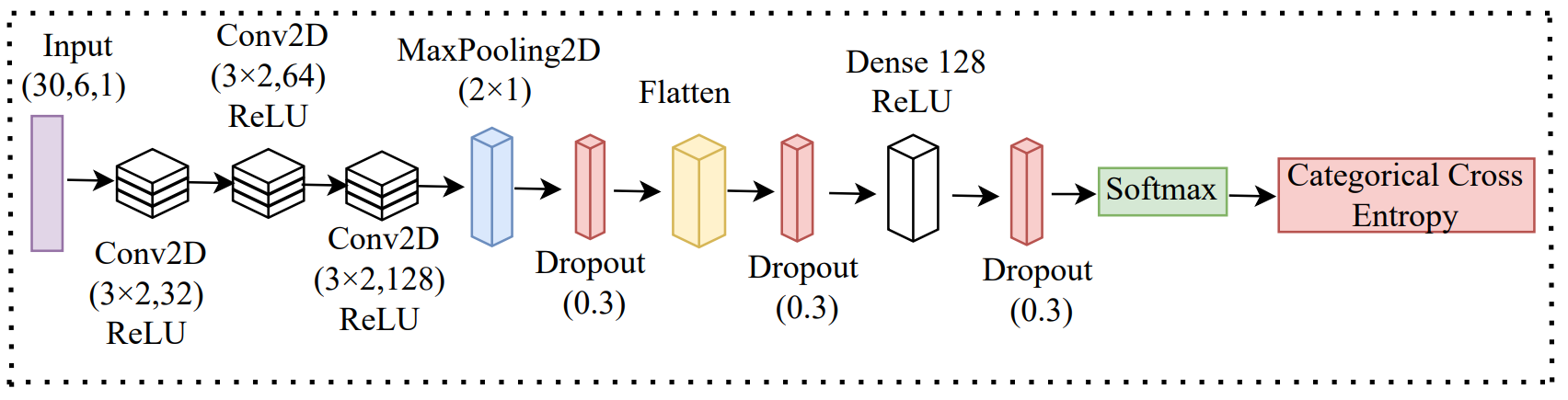}}
\caption{Gesture classification CNN-based model.}
\label{gesture_detetion_model}
\end{figure*}

Previously, researchers have implemented CNNs in classifying gestures from IMU data \cite{zhang2023multistream} \cite{kimGesture}. Depending on previous research to run the model efficiently in DPU, we choose CNN to classify gestures. The proposed gesture classification model Figure \ref{gesture_detetion_model} is specifically designed for efficient deployment on a KV260 DPU, suitable for resource-constrained environments. It comprises three convolution layers (with 32, 64, and 128 filters, respectively), a MaxPooling2D, Flatten, and followed by a Dense layer with 128 neurons. A Dropout layer after Maxpooling2D , Flatten and Dense layer was added.  The use of relatively small convolutional kernels (5×2) and (3x2) pooling (2×1)  reduces computational load and memory requirements. However, the simplified architecture, while optimized for low-power execution, may lead to limited capability in extracting intricate spatial-temporal features, potentially impacting performance on more complex gesture datasets. Despite this limitation, the model presents a well-balanced compromise between computational efficiency and accuracy, especially suited for real-time gesture recognition tasks on edge devices like the KV260.

For comparison, the Long Short-Term Memory (LSTM) and Gated Recurrent Unit (GRU) models were also trained with the same dataset and epochs; both LSTM and GRU models have a two-layer LSTM or GRU network with 64 and 32 units, followed by dropout regularization, a dense ReLU layer and a softmax output layer, designed for multiclass gesture classification.

\subsubsection{Gesture data collection and training}
\label{sec:gestureData}
The raw IMU data were collected using our developed hand using I2C. A button is used to start the data sampling process, thereby reducing the influence of unintentional hand movements. The button is activated at the time of the initiation of the gesture, and data collection stops after an exact number of samples is collected. We have collected total data for three hand movements (Tilt right and Tilt left, No action). We have collected 120 samples for each gesture at 30 Hz. The total collected dataset consists of 660 IMU samples at a sampling rate of 30 Hz, categorized into three gesture classes: twist right, twist left, and no action. The data was distributed into training (80\%), testing (10\%), and validation (10\%) sets, and trained for 300 epochs with a learning rate of 0.001 and a batch size of 32.

\subsection{Custom PCB design}
\begin{figure}[!h]
\centering
\includegraphics[width=\columnwidth]{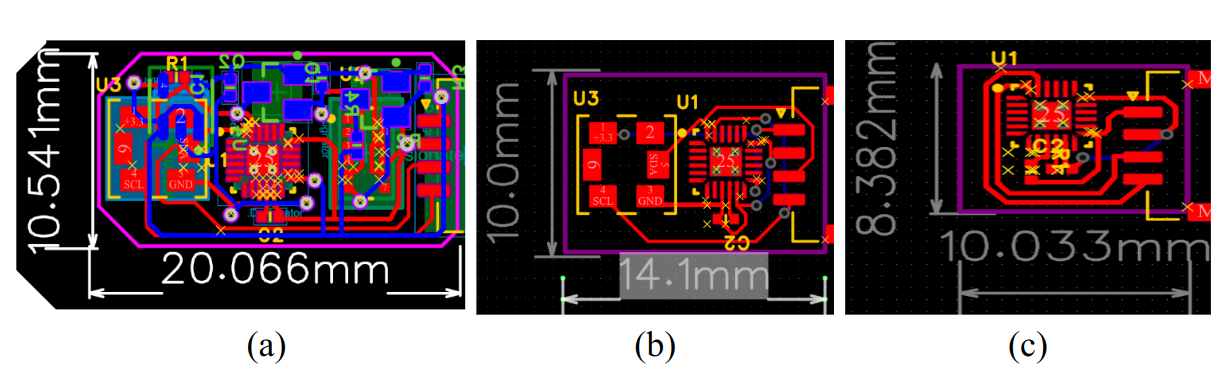}
\caption{Designed PCBs for placing sensors on the finger. (a) PCB design for index finger fingertip. (b) PCB design for distal phalanx of other fingers (c) PCB design  for all five finger middle phalanx}
\label{designed_pcbs}
\end{figure}

Printed circuit boards (PCBs) were designed and printed, illustrated in Figure \ref{designed_pcbs}. Due to limited space at the fingertips, designing these PCBs posed considerable challenges related to component density, routing complexity, and overall board size constraints. With dimensions around 10–20 mm, enabling precise placement within small prosthetic finger segments. However, the complexity of miniaturization and dense routing increased signal interference, and thermal issues were considered while designing. 

\subsection{Hosting the models on FPGA}
A custom hardware overlay was created to deploy DL models on the Xilinx UltraScale+ FPGA (on the dev board KV260) with custom peripherals via PYNQ. We designed hardware architecture in Vivado 2023, defining interfaces for peripherals (e.g., cameras, sensors, motors) and integrating the Deep Learning (DL) Processing Unit (DPU) IP core. After designing the overlays, we generated bitstream and then transferred it to KV260. This overlay is loaded onto the KV260 FPGA using PYNQ APIs, allowing Python code to interface with both the DL accelerator and your custom peripherals. We followed the standard Vitis AI procedures for model training, quantization to INT8 precision, and deployment and hosted the quantized model on the FPGA using the DPU runtime provided within the PYNQ environment.

The project environment for training and deploying an object detection model on an FPGA is trained on a workstation with an Intel Core i9 processor, 64GB of RAM, and an RTX 3090 Ti GPU. The software setup uses Ubuntu 20.04 LTS with Vitis AI Docker image version 3.5, while FPGA overlay development is managed with Vivado 2023. The hardware design workflow includes PCB design with EasyEDA Pro and 3D modeling using Fusion 360.

\subsection{Power Management}
The prosthetic hand is powered by an 11.1V 1300mAh LiPo battery connected to a buck converter that provides both 3.3V and 2.8V via separate voltage regulators. Two level shifters are used to accommodate the TOF (VL6180) distance sensor, which operates at 2.8V, while another sensor on the same bus requires 3.3V.

\section{Result}
\label{sec:result}
\subsection{Object detection Result}
The trained YOLOv7-tiny QAT model achieved a mean average precision (mAP@0.5) of ~96\% for both training and testing datasets. The system was implemented using a PYNQ board and a USB camera to capture real-time video feeds. Real-time processing was performed on DPU; it was able to perform nine frames per Second (FPS).

\subsection{Fore and Grasp Classification Result}
The model achieved a grasp selection accuracy of 100\% in classifying the three grasping patterns (power grip, pinch, pronated) and the force estimation Mean Absolute Error (MAE) of 0.0181.

\subsection{Gesture classification Result}
The Gesture classification CNN model achieved a training accuracy of 99\% and a testing accuracy of 100\%. The comparison of different models trained with the dataset is shown in Table \ref{tab:model_performance}.

\begin{table}[h]
\centering
\caption{Model Performance Comparison for gesture classification}
\begin{tabular}{lcc}
\hline
\textbf{Model} & \textbf{Training Accuracy} & \textbf{Test Accuracy} \\ \hline
LSTM & 96\%   & 97\% \\  
GRU & 99\%   & 99\% \\  
\textbf{Our Model (CNN)} & \textbf{99\%}   & \textbf{100\%} \\  
\hline
\end{tabular}
\label{tab:model_performance}
\end{table}


\section{Discussion}

The innovation of this developed system lies in its unique approach to implementing custom DL models on an FPGA for real-time image processing and adaptive grasping guidelines. This design addresses the typical limitations of AI systems by enabling high-performance functionality on resource-constrained embedded FPGA platforms, eliminating the need for expensive GPU-based processors. 

This work demonstrates a systematic integration of vision and embedded computing within a prosthetic hand designed for children with upper limb disabilities by deploying a quantized YOLOv7-tiny model on an FPGA-based Deep Learning Processing Unit (DPU) while processing nine frames per second. These performance metrics indicate that the approach is viable for real-time object detection under controlled conditions.

We addressed the limitations of commercially available EMG/EEG-based systems by developing a custom camera-based prosthesis with novel interfacing hardware, creating a lightweight design. The "hand shell" (from the wrist to the fingers) weighs 105 gm; without the battery and arm shell, the complete hand weighs approximately 400 gm.. The firmware for low-power embedded processors presents a pipeline for prosthetic hands to run in low-resource environments. This firmware can be easily adapted for other prosthetic applications and robotic research, expanding its potential reach. Importantly, this system is designed to be low-cost, making it accessible to people in developing countries, who often lack access to advanced prosthetic technologies. Our designed prosthetic hand delivers a finger force of up to 5 N to lift a 500 ml bottle, with a closing and opening time of approximately 1.5 seconds and 0.6 seconds, respectively. This finger force can be increased by implementing more robust servo motors. The closing time is a little longer than the opening time as we are measuring force continuously while closing the hand for grasping. 

By integrating low-cost AI-vision technology into powered prosthetic hands, the system bypasses expensive proprietary control electronics and interfaces, directly connecting with the motors of the hand.  The ability to select features and sensors based on individual requirements enables the creation of energy-efficient, affordable designs for users who may not need the full range of functionalities. This flexibility makes the system not only cost-effective but also accessible to populations in the developing world, where affordability and simplicity are critical. 

One key advantage of implementing all control and processing logic in an FPGA is the potential to transition to a dedicated ASIC design for mass production. By first validating the system’s functionality and performance in the flexible FPGA environment, developers can confidently move to an  Application-Specific Integrated Circuit (ASIC) solution that offers lower unit costs, reduced power consumption, and compact form factors—ideal for high-volume manufacturing scenarios.

The decision to use a camera-based method for control, as opposed to EMG or EEG-based approaches, helps reduce the need for extensive user-specific training and calibration. Instead, the prosthesis leverages sensor fusion, combining data from a 2MP USB camera, pressure sensors, and IMUs to reliably classify grasp patterns and control motor functions. The integration of a CNN-based gesture recognition model was driven by compatibility constraints with the KV260 board. Although alternative architectures such as LSTM and GRU showed marginally higher performance, the chosen CNN architecture offered a better fit for the system’s limited resources and provided acceptable accuracy levels.

This machine-vision research had immense societal importance, as children in early and middle childhood with limb loss were under-served by current prosthetic hand options. The constant growth of children requires frequent replacement of their prostheses; this prosthetic hand electronics is designed in such a way it should fit with the larger hand, which will reduce the complexity and cost. 

In this design, we made all the finger sizes the same. However, we introduced an angle difference in the joints to make it more humanoid. Making each finger size mimic the human hand would be a more robust solution. While applying silicone; the ratio was determined by empirically different combinations of Silicone base and curing agent to get the right rigidness. From a mechanical perspective, the underactuated design strikes a balance between achieving sufficient dexterity and maintaining simplicity, low weight, and cost-effectiveness. However, further standardization and testing are needed, especially for long-term use and durability. For communication, we selected I2C for the finger module because SPI would need additional wiring, and the child's hand’s finger design has very limited space.

Convolutional Neural Networks (CNNs) offer distinct advantages over traditional fully connected (Dense) layers when deploying gesture recognition models on the KV260 DPU. CNNs use weight-sharing and local receptive fields, significantly reducing the number of parameters, computational complexity, and memory footprint compared to Dense layers, thus enhancing hardware efficiency. This optimization leads to faster inference, lower latency, and better suitability for real-time gesture recognition tasks on resource-constrained embedded systems. Additionally, CNN architectures naturally extract spatial and temporal features from IMU sensor data, providing improved accuracy over Dense layers, which typically treat each input dimension independently. Consequently, CNN-based models are inherently more scalable and performant, making them ideal candidates for deployment on the KV260 DPU platform. 

In our Gesture Recognition model, we initially started with a small CNN. At first, we did not include a dropout layer, which resulted in high validation and test loss. After adding a dropout layer, the loss decreased. While testing the model, we achieved 100\% accuracy; however, we only had 20 data points. More data collection and further testing are required to validate the model's accuracy for diverse environment.

Despite these promising results, there remain several areas for improvement. One notable limitation is the absence of slip detection capabilities, which could enhance grasp stability in dynamic scenarios. Additionally, the current dataset for object detection and gesture recognition may not sufficiently capture the variability encountered in everyday environments. Expanding the training dataset to include a wider range of lighting conditions, backgrounds, and object classes will be necessary to improve the model's generalizability.

\section{Conclusion}
\label{sec:conclusion}
The developed prosthetic hand successfully integrates advanced vision, embedded computing, and 3D printing technologies, offering a functional and cost-effective alternative for children aged 10–12 with upper limb disabilities. Using FPGA-based hardware, real-time object detection with YOLOv7 tiny, and customized grasp classification, the system achieved good performance in both grasping accuracy and gesture recognition. However, the current design does not incorporate slip detection, which could limit its capability in dynamically adjusting grip force. Future work includes integrating slip detection sensors to enhance grasp stability, expanding the object detection dataset with diverse daily-life items under varying conditions, and further refining the CNN-based gesture recognition model for improved accuracy and responsiveness on edge computing platforms. By offering an adaptable, user-friendly, and low-cost solution, the project has the potential to significantly improve the quality of life for individuals in third-world countries, bringing advanced prosthetic technology within reach of those who previously could not afford it.

\section{Acknowledgment}
We would like to extend our gratitude to Ernesto Sola-Thomas, Juan Publo Sola-Thomas, and Salwa Omar from Clarkson University for their invaluable support and guidance at various stages of this work. Their expertise and willingness to help have greatly enriched this project, and we are deeply appreciative of their contributions.

\bibliographystyle{IEEEtran} 
\bibliography{bibtext}

\begin{IEEEbiography}[{\includegraphics[width=1in,height=1.25in, clip, keepaspectratio]{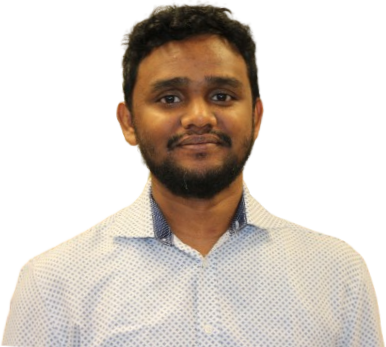}}]{MD ABDUL BASET SARKER} is currently a PhD candidate in the Department of Electrical and Computer Engineering at Clarkson University, Potsdam, NY, USA, and an active researcher at the AI Vision, Health, Biometrics, and Applied Computing (AVHBAC) Lab. He specializes in Artificial Intelligence and Machine Vision, with his research primarily focused on developing assistive technologies such as vision-enabled prostheses, exoskeletons, and AI-powered systems designed for low-power devices. He earned his Bachelor's degree in Electronics and Communication Engineering from the National University of Bangladesh and subsequently pursued his Master’s degree, followed by ongoing PhD studies at Clarkson University. His projects involve edge computing, embedded systems integration, deep learning, and SLAM technologies, targeting applications in healthcare, environmental monitoring, and autonomous navigation.

\end{IEEEbiography}

\begin{IEEEbiography}[{\includegraphics[width=1in,height=1.25in, clip, keepaspectratio]{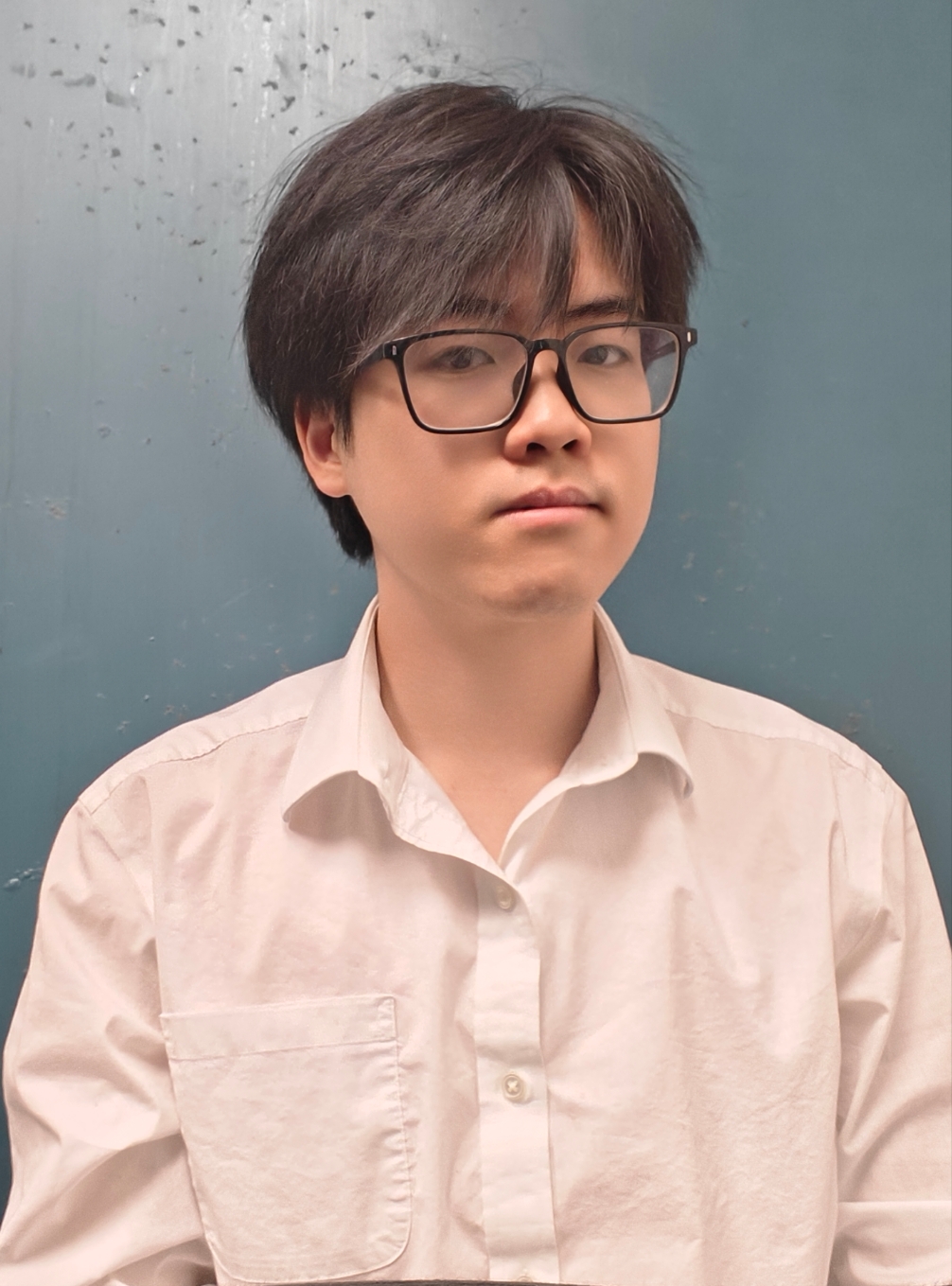}}]{ART NGUYEN}  is currently an undergraduate student in the Department of Electrical and Computer Engineering at Clarkson University, Potsdam, NY, USA where he is actively engaged in research in the fields of AI and embedded systems. He has been working as an assistant researcher in the AI Vision, Health, Biometrics, and Applied Computing (AVHBAC) lab at Clarkson University.

\end{IEEEbiography}

\begin{IEEEbiography}[{\includegraphics[width=1in,height=1.25in, clip, keepaspectratio]{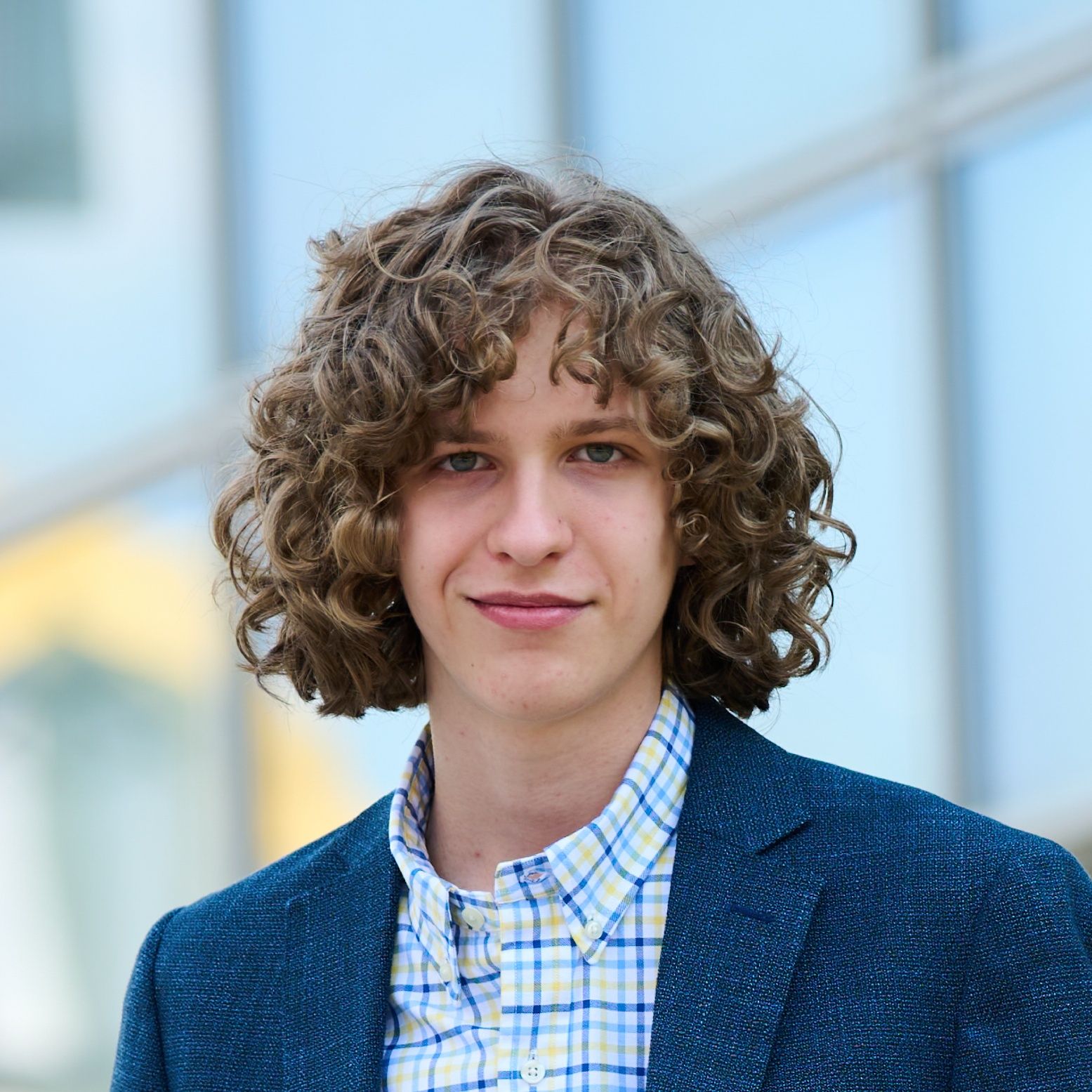}}]{SIGMOND KUKLA} is an undergraduate student at Clarkson University, class of 2028, pursuing a double major in Electrical and Computer Engineering. He is an Ignite Presidential Fellow as well as a member of the Honors Program. Sigmond is actively involved in research as an Undergraduate Research Assistant within Clarkson's Center for Advanced PCB Design and Manufacturing (CAPDM) and AI Vision, Health, Biometrics, and Applied Computing (AVHBAC) labs. He has contributed to a number of projects in the sensor and embedded systems design space. Sigmond is also an entrepreneur within the VR game development field, which has also been applicable to his recent research, including balance testing and gaze detection.

\end{IEEEbiography}

\begin{IEEEbiography}[{\includegraphics[width=1in,height=1.25in, clip, keepaspectratio]{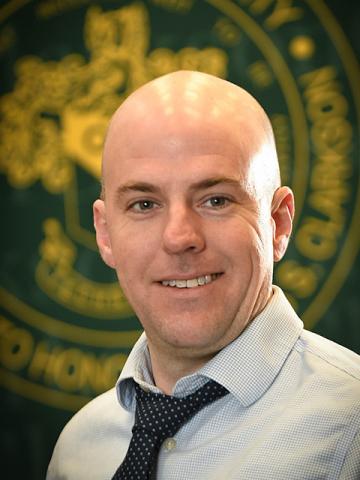}}]{KEVIN FITE} 
Kevin Fite is an Associate Professor in the Department of Mechanical and Aerospace Engineering, Clarkson University, Potsdam, New York.  He received the B.E., M.S., and Ph.D. degrees in mechanical engineering from Vanderbilt University in 1997, 1999, and 2002, respectively. From 2002 to 2007, he was a Research Associate in the Department of Mechanical Engineering, Vanderbilt University.  He joined Clarkson University in 2007.  Research interests include the design and control of electromechanical and fluid power systems with application in assistive and rehabilitation technology.  

\end{IEEEbiography}

\begin{IEEEbiography}[{\includegraphics[width=1in,height=1.25in, clip, keepaspectratio]{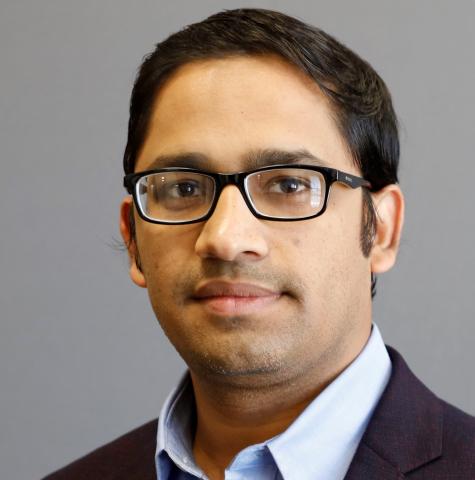}}]{MASUDUL H.
IMTIAZ} is currently an assistant professor with the Department of Electrical and Computer Engineering, Clarkson University, Potsdam, NY, USA, and head of the AI Vision, Health, Biometrics, and Applied Computing (AVHBAC) lab. Dr. Imtiaz received bachelor’s and master’s degrees in applied physics, electronics, and communication engineering from the University of Dhaka, Bangladesh, and a Ph.D. degree from the University of Alabama in the summer of 2019. He was a Postdoctoral Fellow with the Department of Electrical and Computer Engineering at the University of Alabama. His research interests include the development of wearable systems, mHealth, deep learning on wearables, biomedical signal processing, and computational intelligence for preventive, diagnostic, and assistive technology.

\end{IEEEbiography}

\EOD

\end{document}